\documentclass[11pt]{article}

\PassOptionsToPackage{sort}{natbib}
\usepackage[preprint]{acl}

\usepackage{times}
\usepackage{latexsym}

\usepackage[T1]{fontenc}

\usepackage[utf8]{inputenc}

\usepackage{microtype}

\usepackage{inconsolata}

\usepackage{graphicx}
\usepackage{orcidlink}

\usepackage{caption}
\usepackage{algorithm}
\usepackage{algorithmic}
\usepackage{adjustbox}
\usepackage{multirow}
\usepackage{listings}
\usepackage[most]{tcolorbox}
\usepackage{bm}
\usepackage{subcaption}

\usepackage{xspace}
\usepackage{booktabs}
\usepackage{float}

\usepackage{pgfplots}
\usepgfplotslibrary{groupplots}
\usetikzlibrary{patterns, calc}
\pgfplotsset{compat=1.18}

\usepackage{colortbl}
\usepackage{newfloat}
\usepackage{amsmath, mathtools}
\usepackage[capitalise,nameinlink]{cleveref}

\usepackage{amsfonts}

\crefname{figure}{Fig.}{Figs.}
\Crefname{figure}{Fig.}{Figs.}
\crefname{subfigure}{Fig.}{Figs.}
\Crefname{subfigure}{Fig.}{Figs.}
\crefname{table}{Tab.}{Tabs.}
\Crefname{table}{Tab.}{Tabs.}
\crefname{section}{Sec.}{Secs.}
\Crefname{section}{Sec.}{Secs.}
\crefname{subsection}{Sec.}{Secs.}
\Crefname{subsection}{Sec.}{Secs.}
\crefname{equation}{Eq.}{Eqs.}
\Crefname{equation}{Eq.}{Eqs.}
\crefname{algorithm}{Alg.}{Algs.}
\Crefname{algorithm}{Alg.}{Algs.}
\crefname{appendix}{App.}{Apps.}

\title{Grounding Free-Form Instructions for\\Fashion Complementary Image Generation}

\author{
  {\bfseries
  Matteo Attimonelli$^{1,2}$,
  Claudio Pomo$^{1}$,
  Alessandro De Bellis$^{1}$,
  Danilo Danese$^{1}$} \\
  {\bfseries
  Dietmar Jannach$^{3}$,
  Tommaso Di Noia$^{1}$} \\
  \\
  $^{1}$Politecnico di Bari, Italy \quad
  $^{2}$Sapienza University of Rome, Italy \\
  $^{3}$University of Klagenfurt, Austria
}

\def\styleflow{StyleFlow\xspace}
\def\geco{GeCo\xspace}
\def\mgcm{MGCM\xspace}
\def\ptp{Pix2PixCM\xspace}
\def\fttb{FashionTaobao-TB\xspace}

\begin{document}
\maketitle

\begin{abstract}
    Fashion complementary image generation (CIG) aims to create garments that stylistically match a seed item based on user intent, making it a natural multimodal grounding problem where models must interpret language in visual context.
    Existing CIG benchmarks rely on rigid template prompts (e.g., “a photo of a skirt”), failing to reflect natural user queries and obscuring model behavior across levels of linguistic specificity.
    We introduce \emph{fashion complementary image generation with free-form instructions}, a multimodal language-grounding setting where a model generates a compatible garment from a seed image and a natural-language instruction. To this end, we enrich three CIG benchmarks with low-, medium-, and high-specificity instructions generated by a vision-language model and validated by human annotators.
    We instantiate the task with \emph{\styleflow}, a Rectified Flow Matching model that jointly conditions on the seed image and instruction within a single multimodal transformer.
    Across image quality metrics, catalog-alignment analysis,
    ablations, and human evaluation, \emph{\styleflow} consistently
    produces instruction-aligned and stylistically coherent garments
    while reducing architectural complexity and inference cost
    relative to auxiliary-module approaches.
\end{abstract}

\section{Introduction}

\begin{figure}[t]
\centering
\includegraphics[width=\linewidth]{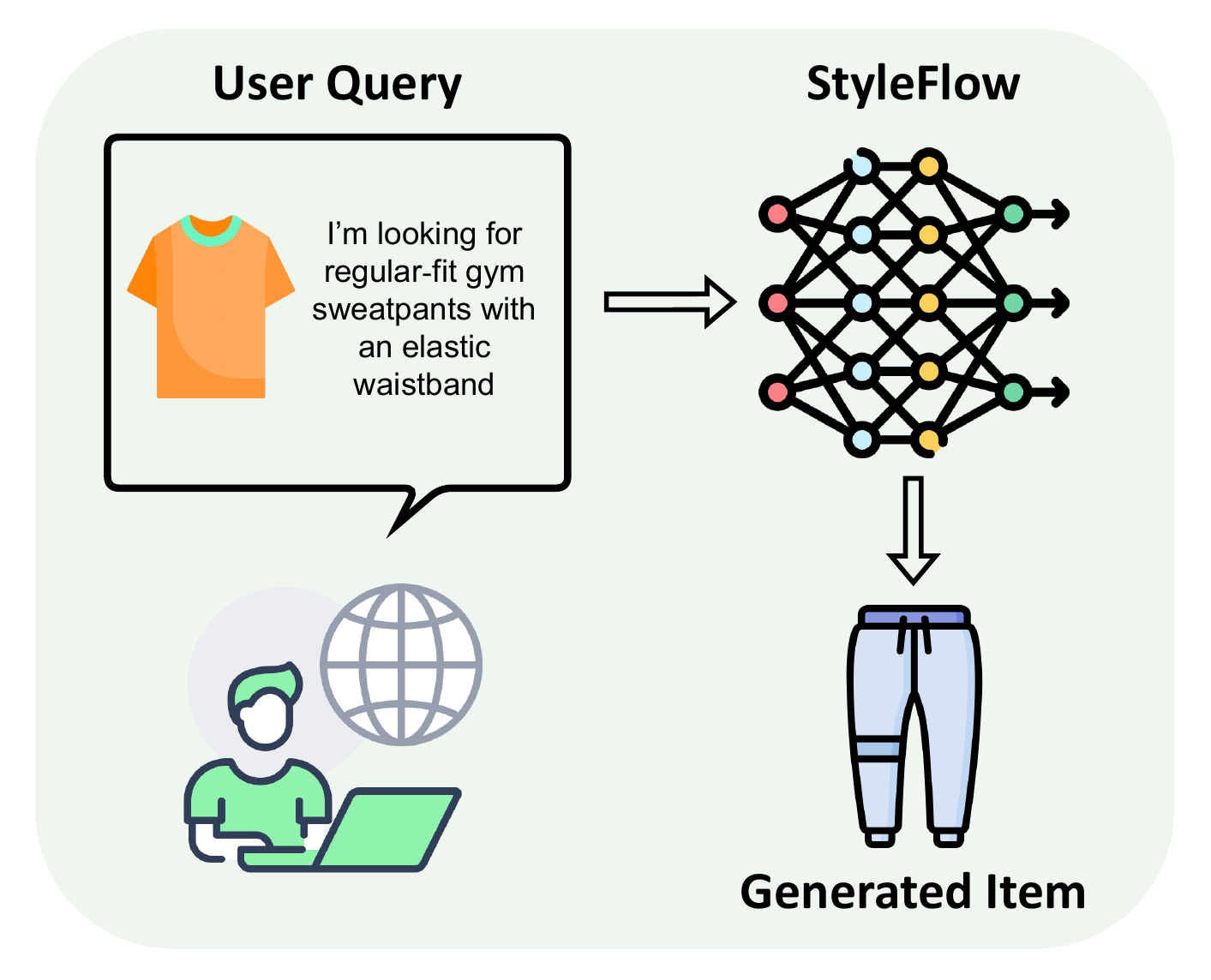}
\caption{The proposed interaction paradigm. Given a seed garment and a free-form instruction, the system generates a complementary product image.}
\label{fig:styleflow_workflow}
\end{figure}

Fashion complementary image generation (CIG) aims to synthesize a garment that stylistically complements a given seed item~\cite{DBLP:conf/ijcai/LiuSRNT020,DBLP:journals/corr/abs-2408-09847,DBLP:journals/ijon/LiuSCM20}, supporting applications such as personalized recommendation~\cite{DBLP:journals/csur/DeldjooNRMPBN24}, design ideation~\cite{DBLP:journals/corr/abs-2306-05182,DBLP:conf/aaai/XieLDLDC25}, and complementary item retrieval~\cite{DBLP:conf/www/BibasSJ23}.
Unlike image editing~\cite{DBLP:conf/iccv/BaldratiMCC0C23} or virtual try-on~\cite{DBLP:conf/cvpr/ChoiPLC21,DBLP:conf/mm/MorelliBCC0C23}, which primarily perform localized or structure-preserving transformations, CIG requires generating an item from a different garment category that remains compatible with the seed. This makes CIG a natural testbed for multimodal language grounding: a model must interpret textual intent while using the seed image as visual context for style, color, and compatibility.

In fashion search and recommendation scenarios, users may express preferences through free-form natural language~\cite{DBLP:journals/entcom/MallikTC25}. Queries may vary substantially in specificity, from underspecified cues such as ``jeans'', to medium-detail requests such as ``blue wide-leg jeans'', to detailed descriptions including material, fit, and visual details. Still, existing CIG methods are both trained and
evaluated with fixed category templates (e.g., ``a photo of
\textit{[category]}''), so they never learn to ground free-form
user intent, and current benchmarks provide no signal as to
whether they can.

Early CIG approaches relied on GAN-based architectures~\cite{DBLP:journals/ijon/LiuSCM20,DBLP:conf/ijcai/LiuSRNT020,DBLP:journals/corr/abs-2408-09847}, while recent latent diffusion models improve visual fidelity~\cite{DBLP:conf/sigir/XuWFMZ024,DBLP:conf/sigir/YuM00L025}.
As most LDMs are pretrained for text-to-image synthesis, prior approaches add auxiliary conditioning modules to inject seed-image information~\cite{DBLP:conf/iccv/ZhangRA23,DBLP:conf/aaai/MouWXW0QS24,DBLP:conf/cvpr/BrooksHE23}. While effective, this design increases architectural complexity and memory overhead.

We introduce \emph{fashion complementary image generation with free-form instructions} (\cref{fig:styleflow_workflow}), where the input is a seed garment image paired with a natural-language instruction describing the desired complementary item.
To approximate realistic interaction while preserving experimental
control, we enrich three CIG benchmarks with instructions at three
specificity levels (low, medium, high), generated by a VLM and
validated by human annotators for fluency, informativeness, and
visual grounding.

We instantiate this setting with \textbf{\styleflow}, a Rectified Flow Matching (RFM) model~\cite{DBLP:conf/iclr/LipmanCBNL23,DBLP:conf/iclr/LiuG023} that jointly conditions on a seed image and a free-form instruction within a single transformer, without auxiliary adapters or control branches. This unified design requires only 20 inference steps, simplifying generation relative to prior
auxiliary-module approaches, and enables direct ablation of text and visual inputs without architectural confounds, allowing us to isolate how instruction specificity and seed-image context each affect generation quality. To the best of our knowledge, this is the first application of RFM to CIG.

We evaluate instruction-driven CIG through image quality metrics, catalog-alignment analysis, ablations over language and visual conditioning, and two human studies. Across these evaluations, \styleflow~consistently produces instruction-aligned and stylistically coherent garments while reducing architectural complexity and inference cost relative to auxiliary-module approaches.

Our key contributions are:

\begin{enumerate}
    \item We formalize \emph{fashion CIG with free-form instructions},
    reframing CIG from fixed category-template generation to
    multimodal grounding under varying levels of linguistic specificity.

    \item We enrich three CIG benchmarks with low, medium, and high specificity instructions, human-validated for fluency and visual grounding, enabling controlled evaluation of instruction-driven generation.

    \item We propose \emph{StyleFlow}, a Rectified Flow Matching
    model that integrates seed image and free-form instruction
    conditioning within a single multimodal transformer, eliminating
    auxiliary modules.

    \item We provide a systematic empirical study of how instruction specificity affects generation quality and catalog alignment, combining automated metrics with human judgments of realism, compatibility, and authenticity.
\end{enumerate}

Together, these contributions move CIG closer to realistic and user-aligned deployment in interactive fashion systems.

\section{Proposed Approach}
\label{sec:methodology}
This section formalizes fashion CIG with free-form instructions as a multimodal grounding task.
We first define the input-output setting and describe how standard CIG benchmarks are enriched with instructions at controlled levels of semantic specificity.
We then present \textit{\styleflow}, an RFM-based model that instantiates this setting by jointly conditioning generation on a seed garment image and a free-form instruction.

\subsection{Problem Formulation}

\emph{Fashion complementary generation with free-form instructions} aims to synthesize a garment that stylistically complements a reference item under a textual user query.
Let $\mathcal{D}$ denote a dataset of compatible garment pairs $(s_i, b_j)$, where $s_i \in \mathcal{S}$ and $b_j \in \mathcal{B}$ represent seed and target categories, respectively.
Each item is associated with an image $I_{s_i}$ or $I_{b_j}$.
Given a seed image $I_{s_i}$ and a free-form instruction $c_{b_j}$ describing the desired complementary item, the objective is to generate an image $\hat{I}_{b_j}$ that is stylistically compatible with $I_{s_i}$ and semantically aligned with $c_{b_j}$.
The instruction may be underspecified or detailed, so the model must ground language and visual context jointly: the seed image provides compatibility cues, while the instruction constrains the target garment attributes.
To study this interaction, existing CIG datasets are enriched with multi-granularity natural language instructions that expose varying levels of specificity.

\subsection{Free-Form Instruction Synthesis}\label{subsec:instruction_generation}

Instructions are constructed to approximate user intent across varying levels of specificity while keeping the benchmark controlled.
Prompts range from minimal (e.g., ``An image of denim jeans'') to detailed, style-rich descriptions
(e.g., ``High-waisted straight-leg blue denim jeans with patchwork panels and a raw frayed hem'').
Each instruction is paired with a seed garment \( s_i \) and used to guide the generation of a compatible target \( b_j \).

The enrichment process follows a two-stage pipeline, implemented using a VLM (see \cref{fig:styleflow_architecture}a).
First, for each target item \( b_j \), the VLM is prompted to generate a structured caption describing key visual attributes such as garment category, color, fit, material, and style.
Then, the same model is prompted to rewrite these captions into natural, free-form instructions.
Each \( b_j \) is associated with three instructions reflecting increasing levels of specificity: (i) \textit{Low} detail prompts mention only the garment category; (ii) \textit{Medium} detail prompts also include category and color; (iii) \textit{High} detail incorporate stylistic aspects such as fit, material, and stylistic cues.
This formulation enables controlled experimentation across a spectrum of instruction complexity, making prompt specificity an explicit evaluation variable.
All generated instructions are manually verified for clarity, semantic relevance, and alignment with the visual content of \( b_j \).
Additionally, a human evaluation (\cref{subsec:human_eval}) assesses the realism and usefulness of these instructions in guiding generation.
To support generalization, test sets are constructed such that no target garment \( b_j \) appears in training, even under different seed pairings \( s_i \).
This design ensures rigorous evaluation of CIG, requiring models to produce outputs for previously unseen items and instructions.
Examples of generated free-form instructions are shown in \cref{subsec:qualitative}, with further details provided in~\cref{app:data_constr}.

\subsection{StyleFlow}
\label{sec:styleflow-model}

\begin{figure*}[t!]
    \centering
    \includegraphics[width=0.98\linewidth]{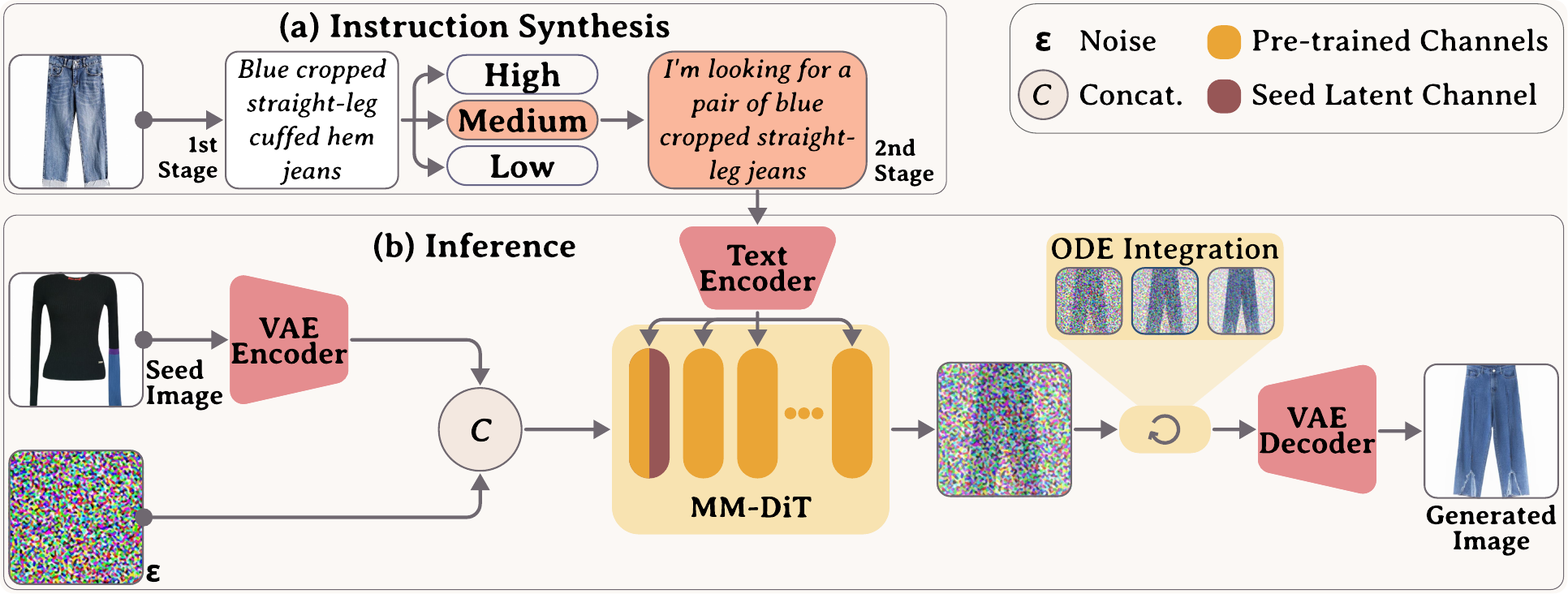}
    \caption{Overview of the method. (a) Illustrates the two-stage instruction synthesis: a VLM generates a caption,  which is then rewritten into prompts at three detail levels. (b) \styleflow inference: the model generates a complementary garment from a seed image and a selected free-form instruction (medium detail shown).}
    \label{fig:styleflow_architecture}
\end{figure*}

\cref{fig:styleflow_architecture}b provides an overview of \styleflow, developed to synthesize a compatible target garment conditioned on a seed image and a free-form instruction. The model builds upon RFM, which uses deterministic flows and provides a convenient backbone for instruction-driven generation.

\paragraph{Rectified Flow Matching.}
\label{sec:preliminaries}
FM is a generative framework that models deterministic flows between a source distribution $p_0$ and a data distribution $p_1$ through a learned time-dependent velocity field $u_t(x)$. Unlike LDMs, which typically solve SDEs, FM integrates an ODE of the form:
\begin{equation}
\frac{d}{dt} \psi_t(x) = u_t(\psi_t(x)), \quad \psi_0(x) = x,
\end{equation}
where $\psi_t$ denotes the flow and $x \sim p_0$. The velocity field $u_t^\theta$ is parameterized by a neural network and trained to transport noise samples $\epsilon \sim \mathcal{N}(0, I)$ toward data samples $x \sim p_1$.
To enable tractable training, RFM is employed. In this setup, a linear path between the source and target distributions $X_t = (1-t)X_0 + t X_1, \quad X_0 \sim \mathcal{N}(0,I),\; X_1 \sim p_1$ is enforced, thereby simplifying the flow dynamic while conditioning on target samples $X_1$, minimizing the following objective:
\begin{equation}
\begin{aligned}
\mathcal{L}_{\mathrm{RFM}}(\theta)
&= \mathbb{E}_{t,X_0,X_1}\!\Big[ \\
&\quad \left\|u_t^\theta(X_t)-(X_1-X_0)\right\|^2
\Big].
\end{aligned}
\end{equation}
This formulation avoids marginalization over intermediate distributions and supports stable conditional generation (further details in~\cref{app:fm_more}).

\sloppy
\paragraph{Complementary Generation.}
StyleFlow builds on FLUX1-Dev~\cite{flux2024}, a high-quality RFM text-to-image model.
FLUX uses a Multimodal Diffusion Transformer (MM-DiT)~\cite{DBLP:conf/iccv/PeeblesX23} to model time-dependent velocity fields in the latent space of a pretrained VAE, originally conditioning solely on textual prompts.
To adapt this framework to instruction-driven CIG, \styleflow introduces joint conditioning on both a seed garment image and a free-form instruction. Following prior work on instruction-driven editing~\cite{DBLP:conf/cvpr/BrooksHE23}, the latent representation of the seed garment \( \tilde{I}_{s_i} \) is concatenated channel-wise with the noise vector \( X_0 \) at the MM-DiT input layer, allowing the model to ground the generated target in the visual context. The initial projection layer of MM-DiT is modified to accommodate the increased channel dimensionality, ensuring compatibility with the pretrained backbone while preserving the overall architectural design.

Free-form instructions \( c_{b_j} \) are encoded in line with the original FLUX~\cite{flux2024} architecture and injected via cross-attention within MM-DiT.
This design gives the model direct access to both sources of grounding: the textual instruction specifies the desired target attributes, while the seed image supplies compatibility context. Despite its simplicity, the joint conditioning design supports generation across garment categories and instruction granularities without the auxiliary fusion networks or complex adapters commonly required in prior LDM-based CIG pipelines~\cite{DBLP:conf/sigir/XuWFMZ024}.

\styleflow is initialized from the FLUX checkpoint, preserving all pretrained weights while randomly initializing the additional seed-latent input channel.
Training is performed via LoRA~\cite{DBLP:conf/iclr/HuSWALWWC22}, applied to the MM-DiT attention blocks, enabling parameter-efficient adaptation while keeping the pretrained backbone frozen and jointly optimizing the newly introduced parameters.

\sloppy
\paragraph{Training Objective.}
Let \( \tilde{I}_{b_j} \) denote the latent of the target garment. Given a noise sample \( X_0 \sim \mathcal{N}(0, I) \), generation is defined via a linear path:
\[
X_t = (1 - t) X_0 + t \tilde{I}_{b_j}, \quad t \in [0, 1].
\]
The model predicts the velocity field conditioned on the seed latent and the instruction, minimizing:
\begin{equation}
\begin{aligned}
\mathcal{L}_{\mathrm{SF}}(\theta)
&= \mathbb{E}_{t,X_0,\tilde{I}_{b_j}}\Big[
\big\|u_t^\theta(X_t\mid \tilde{I}_{s_i},c_{b_j}) \\
&\hspace{4.8em}-(\tilde{I}_{b_j}-X_0)\big\|^2
\Big].
\end{aligned}
\end{equation}
where \( u_t^\theta \) is implemented via the MM-DiT. This setup enables stylistically coherent and instruction-aligned generation across garment categories.

At \textit{inference time}, a noise vector \( X_0 \sim \mathcal{N}(0, I) \) is transported toward the target latent using the learned velocity field \( u_t^\theta \), conditioned on the seed image \( \tilde{I}_{s_i} \) and instruction \( c_{b_j} \).
The flow is integrated over \( t \in [0, 1] \) using a fixed-step ODE solver, yielding a latent sample \( \hat{I}_{b_j} \) that is decoded into an image via the pretrained VAE decoder. Unless otherwise stated, we use 20 integration steps.

\section{Experiments}
\label{sec:experiments}
This section evaluates how free-form instruction specificity affects CIG.
We report image quality and catalog-alignment metrics, ablations over language and seed-image conditioning, and two human studies validating both the generated instructions and the generated garments.

\paragraph{Datasets.}
We evaluate on three top--bottom fashion compatibility benchmarks: \textit{FashionVC}~\cite{DBLP:conf/mm/SongFLLNM17}, \textit{ExpFashion}~\cite{DBLP:journals/tkde/LinRCRMR20}, and \textit{\fttb}~\cite{DBLP:journals/corr/abs-2408-09847}, containing 18,640, 18,640, and 88,326 compatible pairs, respectively.
Following common CIG protocols~\cite{,DBLP:conf/ijcai/LiuSRNT020,DBLP:journals/tkde/LinRCRMR20}, a reduced split of ExpFashion (ExpReduced) is used to ensure controlled comparison across benchmarks with comparable pair distributions, while \fttb is kept at full scale to evaluate larger and higher-resolution conditions.
Each benchmark is enriched with low-, medium-, and high-specificity free-form instructions as described in~\cref{subsec:instruction_generation}.
Instruction synthesis is performed with OpenAI-o3 (API version: 2025-01-01-preview), a VLM independent from the FLUX backbone used by \styleflow.
Bottom garments are disjointly split across train, validation, and test sets to ensure generalization to unseen items.

\begin{figure*}[t!]
    \centering
    \includegraphics[width=0.9\linewidth]{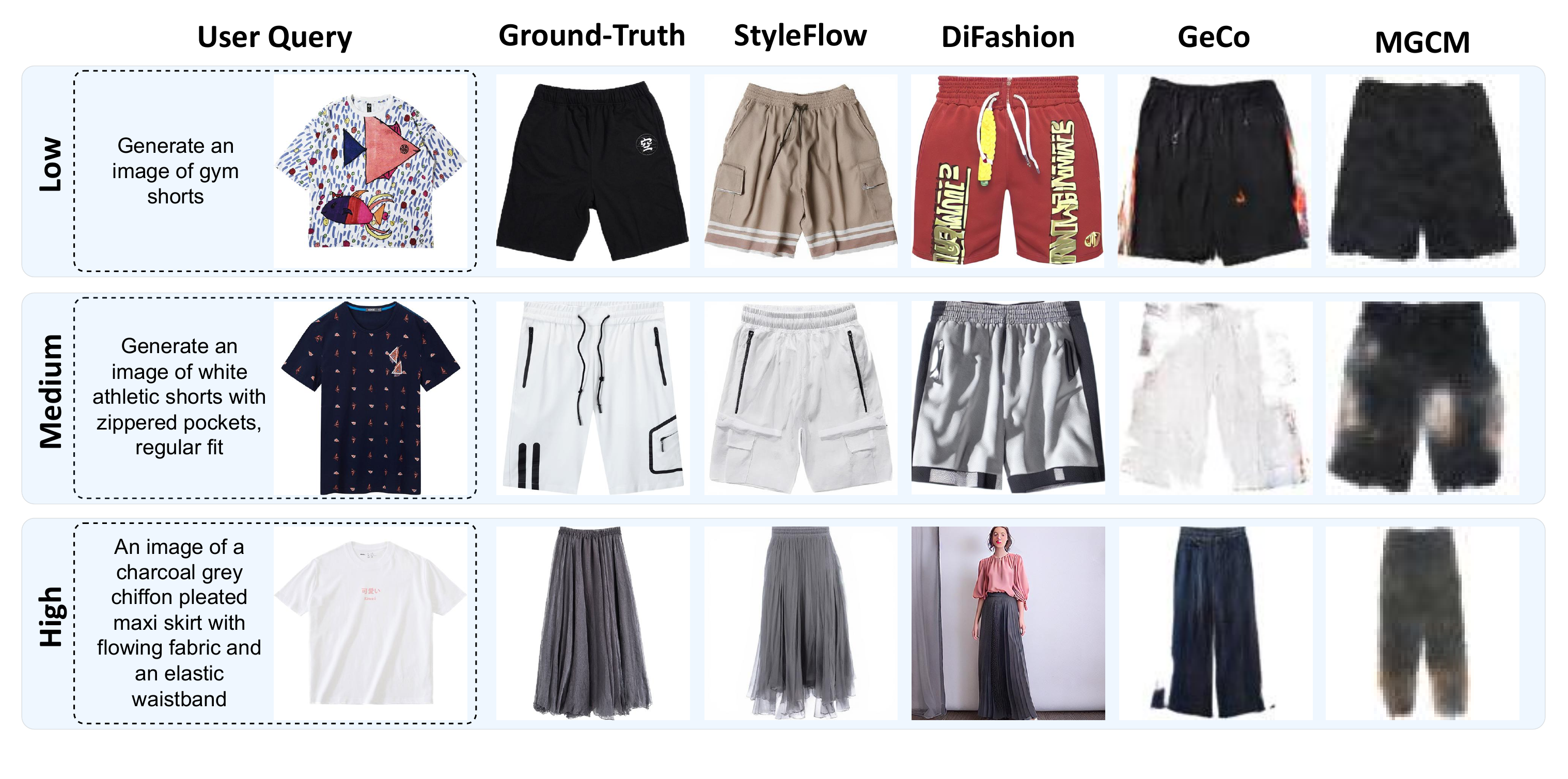}
    \caption{Illustrative samples of bottom garment generation from user queries composed of a top image and free-form instructions at varying levels of detail.}
    \label{fig:styleflow_generations}
\end{figure*}

\paragraph{Evaluation Metrics.}
Generation quality is evaluated with FID, KID, and LPIPS, which measure distributional realism and perceptual similarity to ground-truth garments.
Instruction alignment is assessed with CLIP-Score (CS), which measures consistency between generated images and the input instruction.
For catalog alignment, generated and catalog images are embedded using SigLIP~\cite{DBLP:conf/iccv/ZhaiM0B23} and ranked according to cosine similarity. SigLIP is employed  to mitigate potential architectural bias toward the FLUX backbone.
Alignment is quantified using MRR, Recall, and nDCG, with a cutoff of 50.

\paragraph{Baselines.}
We compare \styleflow with established generative models for fashion CIG. These include \textit{DiFashion}~\cite{DBLP:conf/sigir/XuWFMZ024}, \textit{GeCo}~\cite{DBLP:journals/corr/abs-2408-09847}, \textit{MGCM}~\cite{DBLP:journals/ijon/LiuSCM20}, and \textit{Pix2PixCM}~\cite{DBLP:journals/ijon/LiuSCM20}.
DiFashion is a diffusion-based model operating at $512{\times}512$ resolution, which conditions on top garments and template-based text via a custom module. We adapt it to our setting by conditioning both on the top image and free-form instructions, preserving the original training protocol.
\geco employs a GAN to generate bottom garments from a top garment at $128{\times}128$ resolution, integrating the output into a retrieval pipeline. \mgcm and \ptp build on GANs, generating $64{\times}64$ images used for compatibility-based retrieval. To enable free-form instruction conditioning, both GeCo and MGCM are extended by incorporating a pretrained CLIP text encoder~\cite{DBLP:conf/icml/RadfordKHRGASAM21}, a widely used mechanism to inject textual information.

\paragraph{Implementation Details.}
\styleflow is jointly trained across all datasets for 100{,}000 steps using AdamW~\cite{DBLP:conf/iclr/LoshchilovH19} with cosine learning-rate decay, 100 warm-up steps, bfloat16 precision, and rank-16 LoRA~\cite{DBLP:conf/iclr/HuSWALWWC22}.
The original FLUX weights are frozen, while the LoRA modules and added seed-conditioning input pathway are trained.
Prompt dropout is applied with probability 0.1.
Inference uses classifier-free guidance 3.5 and 20 Euler ODE steps.
Baseline hyperparameters are selected on validation splits following the corresponding implementations. Additional details are provided in~\cref{app:additional_imp_details}.

\subsection{Qualitative Evaluation}\label{subsec:qualitative}
\cref{fig:styleflow_generations} shows generations from \styleflow and baselines across instruction granularities, with additional examples in~\cref{app:more_gen_examples}.
In the \textit{high} setting, DiFashion occasionally introduces extraneous elements (e.g., human body parts), whereas \styleflow generates standalone garments without unintended artifacts or object repetitions.
Under ambiguous \textit{low} prompts, \styleflow often produces items that differ from the ground truth yet remain stylistically compatible, suggesting its usefulness for exploring plausible complementary alternatives and design directions.
These observations are supported by the human evaluation in~\cref{subsec:human_eval}, which assesses realism and instruction alignment.

\begin{table*}[t!]
\caption{Model performance across prompt levels and datasets using FID, KID, LPIPS, and CLIP-Score (CS). Best values in bold, second best underlined.}
\setlength{\tabcolsep}{0.1cm}
\centering
\small
\begin{adjustbox}{width=\textwidth, center} %
\begin{tabular}{@{} c l
                @{\hspace{0.35cm}} cccc
                @{\hspace{0.35cm}} cccc
                @{\hspace{0.35cm}} cccc
                @{\hspace{0.35cm}} @{}}
\toprule
\rotatebox{90}{\textbf{}} & \textbf{Model}
& \multicolumn{4}{c}{\textbf{FashionVC}}
& \multicolumn{4}{c}{\textbf{ExpReduced}}
& \multicolumn{4}{c}{\textbf{FashionTaobao-TB}} \\
\cmidrule(lr){3-6} \cmidrule(lr){7-10} \cmidrule(lr){11-14}
\rotatebox{90}{\textbf{}} & & FID$\downarrow$ & KID$\downarrow$ & LPIPS$\downarrow$ & CS$\uparrow$ & FID$\downarrow$ & KID$\downarrow$ & LPIPS$\downarrow$ & CS$\uparrow$ & FID$\downarrow$ & KID$\downarrow$ & LPIPS$\downarrow$ & CS$\uparrow$ \\
\midrule

\multirow{4}{*}{\rotatebox{90}{High}}
& StyleFlow   & \textbf{33.21}  & \textbf{0.006} & \textbf{0.367} & \textbf{0.319} & \textbf{27.56}  & \textbf{0.003} & \textbf{0.379} & \textbf{0.309} & \textbf{16.93}  & \textbf{0.003} & \textbf{0.335} & \textbf{0.312} \\
& DiFashion   & \underline{67.09}  & \underline{0.026} & 0.622 & \underline{0.291} & \underline{56.62}  & \underline{0.016} & 0.638 & \underline{0.290} & \underline{58.53}  & \underline{0.031} & 0.611 & \underline{0.279} \\
& GeCo        & 97.48  & 0.050 & \underline{0.441} & 0.209 & 99.90  & 0.052 & \underline{0.486} &  0.221 & 60.88  & 0.040 & \underline{0.400} & 0.263 \\
& MGCM        & 182.03 & 0.139 & 0.513 & 0.211 & 218.68 & 0.175 & 0.580 & 0.216 & 165.11 & 0.160 & 0.491 & 0.250 \\
\midrule

\multirow{4}{*}{\rotatebox{90}{Medium}}
& StyleFlow   & \textbf{33.88}  & \textbf{0.006} & \textbf{0.379} & \textbf{0.317} & \textbf{27.56}  & \textbf{0.003} & \textbf{0.389}  & \textbf{0.311} & \textbf{16.78}  & \textbf{0.003} & \textbf{0.340} & \textbf{0.307} \\
& DiFashion   & \underline{73.27}  & \underline{0.029} & 0.623 & \underline{0.289} & \underline{65.02}  & \underline{0.020} & 0.642 & \underline{0.293} & \underline{60.17} & \underline{0.032} & 0.598 & \underline{0.289} \\
& GeCo        & 96.37 & 0.050 & \underline{0.446} & 0.220 & 100.37 & 0.054 & \underline{0.491} & 0.250 & 61.18 & 0.040 & \underline{0.402} & 0.271 \\
& MGCM        & 180.74 & 0.139 & 0.514 & 0.241 & 218.46 & 0.178 & 0.579 & 0.246 & 165.58 & 0.160 & 0.491 & 0.268 \\
\midrule

\multirow{4}{*}{\rotatebox{90}{Low}}
& StyleFlow   & \textbf{46.05}  & \textbf{0.007} & \textbf{0.459} & \textbf{0.289} & \textbf{38.58}  & \textbf{0.005} & \textbf{0.463} & \underline{0.282} & \textbf{22.11}  & \textbf{0.003} & \textbf{0.373} & \textbf{0.294}\\
& DiFashion   & \underline{80.07} & \underline{0.033} & 0.693 & \underline{0.283} & \underline{73.57} & \underline{0.024} & 0.673 & \textbf{0.285} & 72.00 & \underline{0.037} & 0.607 & \underline{0.289} \\
& GeCo        & 109.08 & 0.063 & \underline{0.535} & 0.219 & 111.50 & 0.063 & \underline{0.599} & 0.224 & \underline{66.61} & 0.044 & \underline{0.460} & 0.215 \\
& MGCM        & 171.40 & 0.130 & 0.589 & 0.219 & 207.99 & 0.167  & 0.656 & 0.207 & 167.20 & 0.163 & 0.539 & 0.207 \\
\midrule

\multirow{4}{*}{\rotatebox{90}{Template}}
& StyleFlow   & \textbf{67.91}  & \textbf{0.011} & \textbf{0.491} & \underline{0.238} & \textbf{58.34} & \textbf{0.007} & \textbf{0.498} & \textbf{0.254} & \textbf{27.19}  & \textbf{0.003} & \textbf{0.385} & \textbf{0.279} \\
& DiFashion   & \underline{88.51}  & \underline{0.039} & 0.686 & \textbf{0.261} & \underline{73.82}  & \underline{0.025} & 0.685 & \underline{0.244} & \underline{64.02}  & \underline{0.031} & 0.632 & \underline{0.263} \\
& GeCo        & 104.07 & 0.058 & \underline{0.497} & 0.201 & 107.62 & 0.061 & \underline{0.543} & 0.236 & 67.07 & 0.040 & \underline{0.430} & 0.224 \\
& MGCM        & 178.40  & 0.137 & 0.556 & 0.202 & 214.79 & 0.174 & 0.619 & 0.224 & 164.25 & 0.158 & 0.515 & 0.257 \\
\midrule
\multirow{1}{*}{} & Pix2PixCM & 191.14  & 0.159  & 0.564 & -- & 226.95 & 0.193 & 0.639 & -- & 148.13 & 0.153 & 0.513 & -- \\
\bottomrule
\end{tabular}
\end{adjustbox}
\label{tab:gen_quality}
\end{table*}

\subsection{Quantitative Evaluation}\label{subsec:quantitative}

\paragraph{Generation Quality.}
\cref{tab:gen_quality} shows that instruction specificity is a primary driver of generation quality.
For \styleflow, high- and medium-specificity prompts consistently yield the strongest FID, KID, LPIPS, and CS scores, while low-specificity prompts degrade performance because they provide fewer target attributes.
For example, moving from \textit{low} to \textit{medium}/\textit{high} reduces FID from 46.05 to 33.88/33.21 on FashionVC, from 38.58 to 27.56/27.56 on ExpReduced, and from 22.11 to 16.78/16.93 on \fttb.
The same pattern holds for perceptual similarity and text-image alignment: richer prompts generally reduce LPIPS and improve CS, indicating that additional linguistic detail helps the model resolve target attributes rather than relying only on generic category information.
The structured \textit{template} setting, which approximates prior category-only CIG prompting, is generally weaker than free-form high/medium instructions and narrows the gap between models.
This suggests that the proposed free-form setting exposes a controllable language variable: as instructions become more informative, generation becomes both more realistic and more semantically aligned.
Across datasets and prompt levels, \styleflow is the best or second-best on nearly all metrics, showing that the instruction-driven formulation can be supported without sacrificing visual quality.

\begin{table*}[t]
\caption{Model performance across prompt levels and datasets with MRR, Recall, and nDCG@50. Best values in bold, second best underlined. \rlap{$^{*}$}~~denotes statistically significant improvements over the second best.}
\setlength{\tabcolsep}{0.2cm}
\centering
\small
\begin{adjustbox}{width=\textwidth, center} %
\begin{tabular}{lp{1.5cm}
                ccc
                ccc
                ccc}
\toprule
\textbf{} & \textbf{Model}
& \multicolumn{3}{c}{\textbf{FashionVC}}
& \multicolumn{3}{c}{\textbf{ExpReduced}}
& \multicolumn{3}{c}{\textbf{FashionTaobao-TB}} \\
\cmidrule(lr){3-5} \cmidrule(lr){6-8} \cmidrule(lr){9-11}
\textbf{} & & MRR$\uparrow$  & Recall$\uparrow$ & nDCG$\uparrow$ & MRR$\uparrow$ & Recall$\uparrow$ & nDCG$\uparrow$ & MRR$\uparrow$ & Recall$\uparrow$ & nDCG$\uparrow$ \\
\midrule

\multirow{4}{*}{\rotatebox{90}{High}}
& StyleFlow   & \textbf{0.4811}\rlap{$^{*}$} & \textbf{0.9339}\rlap{$^{*}$} & \textbf{0.5798}\rlap{$^{*}$} & \textbf{0.4745}\rlap{$^{*}$} & \textbf{0.9359}\rlap{$^{*}$} & \textbf{0.5750}\rlap{$^{*}$} & \textbf{0.2189}\rlap{$^{*}$} & \textbf{0.6904}\rlap{$^{*}$} & \textbf{0.3161}\rlap{$^{*}$} \\
& DiFashion   & \underline{0.3223} & \underline{0.8210} & \underline{0.4274} & \underline{0.3459} & \underline{0.7984} & \underline{0.4402} & \underline{0.1240} & \underline{0.5518} & \underline{0.2079} \\
& GeCo        & 0.0188 & 0.2210 & 0.0568 & 0.0111 & 0.1266 & 0.0325 & 0.0361 & 0.2921 & 0.0846 \\
& MGCM        & 0.0182 & 0.1758 & 0.0478 & 0.0144 & 0.1516 & 0.0401 & 0.0318 & 0.2622 & 0.0757 \\
\midrule

\multirow{4}{*}{\rotatebox{90}{Medium}}
& StyleFlow   & \textbf{0.3534}\rlap{$^{*}$} & \textbf{0.8806}\rlap{$^{*}$} & \textbf{0.4648}\rlap{$^{*}$} & \textbf{0.3832}\rlap{$^{*}$} & \textbf{0.9047}\rlap{$^{*}$} & \textbf{0.4931}\rlap{$^{*}$} &\textbf{0.1362}\rlap{$^{*}$} & \textbf{0.6042}\rlap{$^{*}$} & \textbf{0.2303}\rlap{$^{*}$} \\
& DiFashion   & \underline{0.2211} & \underline{0.7048} & \underline{0.3213} & \underline{0.2519} & \underline{0.7469} & \underline{0.3536} & \underline{0.1011} & \underline{0.4757} & \underline{0.1749} \\
& GeCo        & 0.0191 & 0.2161 & 0.0563 & 0.0108 & 0.1203 & 0.0312 & 0.0381 & 0.3034 & 0.0881 \\
& MGCM        & 0.0180 & 0.1806 & 0.0481 & 0.0153 & 0.1578 & 0.0419 & 0.0304 & 0.2597 & 0.0741 \\
\midrule

\multirow{4}{*}{\rotatebox{90}{Low}}
& StyleFlow   & \textbf{0.1048}\rlap{$^{*}$} & \textbf{0.5348}\rlap{$^{*}$} & \textbf{0.1877}\rlap{$^{*}$} & \textbf{0.1116} & \textbf{0.5578}\rlap{$^{*}$} & \textbf{0.1999}\rlap{$^{*}$} & \textbf{0.0568}\rlap{$^{*}$} & \textbf{0.3858}\rlap{$^{*}$} & \textbf{0.1198} \\
& DiFashion   & \underline{0.0878} & \underline{0.4855} & \underline{0.1667} & \underline{0.1009} & \underline{0.5125} & \underline{0.1814} & \underline{0.0502} & \underline{0.3533} & \underline{0.1083} \\
& GeCo        & 0.0178 & 0.1823 & 0.0485 & 0.0109 & 0.1328 & 0.0332 & 0.0235 & 0.1960 & 0.0553 \\
& MGCM        & 0.0115 & 0.1177 & 0.0312 & 0.0084 & 0.1016 & 0.0256 & 0.0266 & 0.2222 & 0.0640 \\
\midrule

\multirow{4}{*}{\rotatebox{90}{Template}}
& StyleFlow   & \textbf{0.0412} & \textbf{0.3048}\rlap{$^{*}$} & \textbf{0.0921}\rlap{$^{*}$} & \textbf{0.0395}\rlap{$^{*}$} & \underline{0.2859} & \underline{0.0863} & \textbf{0.0351}\rlap{$^{*}$} & \textbf{0.2846}\rlap{$^{*}$} & \textbf{0.0789} \\
& DiFashion   & \underline{0.0366} & \underline{0.3000} & \underline{0.0865} & \underline{0.0376} & \textbf{0.2984} & \textbf{0.0869}\rlap{$^{*}$} & \underline{0.0326} & \underline{0.2647} & \underline{0.0766} \\
& GeCo        & 0.0163 & 0.1645 & 0.0442 & 0.0100 & 0.1188 & 0.0302 & 0.0291 & 0.2322 & 0.0670 \\
& MGCM        & 0.0136 & 0.1339 & 0.0358 & 0.0100 & 0.1156 & 0.0295 & 0.0280 & 0.2434 & 0.0691 \\
\midrule
\multirow{2}{*}{}
& Pix2PixCM   & 0.0079 & 0.0952 & 0.0238 & 0.0059 & 0.0891 & 0.0212 & 0.0343 & 0.2472 & 0.0745 \\
\bottomrule
\end{tabular}
\end{adjustbox}
\label{tab:retrieval_results}
\end{table*}

\paragraph{Catalog Alignment.}
\cref{tab:retrieval_results} reports catalog-alignment results.
Across datasets, \styleflow benefits consistently from richer instructions: Recall@50 increases from \textit{low} to \textit{medium} to \textit{high} on FashionVC ($0.5348 \rightarrow 0.8806 \rightarrow 0.9339$), ExpReduced ($0.5578 \rightarrow 0.9047 \rightarrow 0.9359$), and \fttb ($0.3858 \rightarrow 0.6042 \rightarrow 0.6904$), with similar trends for MRR and nDCG.
This indicates that more specific language helps generated garments align more closely with the intended catalog item.
Template prompts perform closer to low-specificity instructions, suggesting that category-only language leaves much of the grounding problem unresolved.
DiFashion also improves with richer prompts but remains below \styleflow in most settings, while GAN-based baselines show weaker sensitivity to instruction specificity.
Overall, \styleflow achieves the strongest catalog alignment in most conditions, with statistically significant improvements over the strongest competing baseline in most settings (Wilcoxon signed-rank test, $p < 0.05$).

\subsection{Human Evaluation}\label{subsec:human_eval}

\paragraph{Free-Form Instructions.}
To assess whether the generated instructions provide plausible language inputs, we conducted a first evaluation with 50 participants recruited among researchers (aged 18--50, balanced gender split) with no compensation.
Participants rated 471 instruction-image pairs (three instructions per 157 bottom garments).
Each instruction was evaluated along two criteria: (Q1) perceived instruction quality (clarity, fluency, informativeness), collected on a 5-point Likert scale with half-star granularity, and (Q2) visual grounding, reported as the percentage of positive judgments that the instruction matched the garment appearance.
More details of the protocol are reported in \cref{app:human_eval_details}.
Results confirm that instruction quality improves with specificity: \textit{high} prompts received the highest average rating ($4.31 \pm 0.86$), followed by \textit{medium} ($3.71 \pm 0.81$), and \textit{low} ($2.90 \pm 1.11$).
A similar trend was observed for visual grounding, with high-detail prompts yielding the best alignment with the garment image ($94 \pm 2.4\%$), followed by medium ($93 \pm 2.5\%$) and low ($80 \pm 3.5\%$).
\emph{These results support the use of the generated prompts as controlled natural-language inputs for studying instruction specificity in CIG.}

\paragraph{Generation Quality and Compatibility.}
A second evaluation was conducted with 50 participants following the same selection protocol to assess the generated garments, covering 1,227 cases.
Each case consisted of a top image and three bottom candidates sampled from StyleFlow, DiFashion, MGCM, Pix2PixCM, or the ground truth.
For each set, participants evaluated: (Q1) visual quality on a 0--5 scale, (Q2) stylistic compatibility with the top on the same 0--5 scale, and (Q3) perceived authenticity through a binary real/generated judgment.
To ensure fair comparison, the three generated candidates for each top were produced using instructions at the same specificity level (e.g., all \textit{medium}), except for Pix2PixCM and the ground truth, which are not conditioned on text.
Specificity levels were balanced across trials, and all candidates were displayed at a uniform on-screen resolution to ensure visual fairness.
Each trial was presented individually and in randomized order to prevent position bias.
StyleFlow achieved the highest visual quality score ($4.38 \pm 0.71$), close to ground-truth items ($4.35 \pm 0.79$) and ahead of DiFashion ($4.08 \pm 0.92$), MGCM ($0.78 \pm 0.49$), and Pix2PixCM ($0.71 \pm 0.36$).
StyleFlow also led in compatibility ($3.24 \pm 1.29$), ahead of DiFashion ($2.56 \pm 1.48$), MGCM ($2.57 \pm 1.29$), and Pix2PixCM ($2.46 \pm 1.27$), and approaching the ground truth ($3.31 \pm 1.35$).
For authenticity, fake-rate is reported, where lower is better: StyleFlow was judged fake in 27.9\% of cases, close to ground truth (25.6\%) and lower than DiFashion (55.9\%), MGCM (67.4\%), and Pix2PixCM (63.2\%).
\emph{These results suggest that StyleFlow produces instruction-aligned garments that are perceptually convincing, compatible with the seed, and difficult to distinguish from real catalog items.}

\subsection{Ablation Study}\label{subsec:ablation}

This section analyzes how seed image and free-form instruction specificity affect image quality and catalog alignment.

\paragraph{Impact of Instructions.}
To assess the role of textual guidance, model performance is compared between the \textit{empty} prompt setting (no instruction) and the \textit{low} detail prompts on \fttb.
Among all models, StyleFlow exhibits the strongest dependence on textual guidance: removing the instructions leads to a substantial performance drop (FID: +189.9\%, Recall: –81.7\%), followed by DiFashion (FID: +147.2\%, Recall: –82.6\%).
These pronounced differences indicate that both models are highly responsive to textual input and rely on it to produce instruction-aligned, high-fidelity outputs. In contrast, GeCo and MGCM exhibit only marginal changes (GeCo: FID +0.06\%, Recall –3.1\%; MGCM: FID +0.02\%, Recall –10.0\%), suggesting a weaker capacity to leverage textual cues.
\emph{These findings show that free-form language is not incidental in the proposed setting: removing the instruction substantially changes generation quality and catalog alignment.}

\paragraph{Impact of Seed Image.}
To determine whether \styleflow leverages visual context, model performance is compared with and without access to the seed top image. In the latter case, the seed is replaced with a blank image. When the seed image is blank, \styleflow shows substantial changes in performance, with FID increasing by 318.2\%, 32.0\%, and 30.4\%, and Recall decreasing by 86.0\%, 4.0\%, and 2.5\% on ExpReduced under low, medium, and high detail prompts, respectively. Similar patterns are observed on \fttb, where FID increases by 332.1\%, 1.7\%, and 1.2\%, and Recall decreases by 77.5\%, 3.7\%, and 2.4\%. \emph{These results demonstrate that the importance of visual context depends on instruction specificity.} In the \textit{low} setting, where language provides little target detail, the seed image is essential; under richer prompts, language supplies more attributes and the relative effect of removing the seed becomes smaller.

\section{Related Work}\label{sec:related_work}

\paragraph{Fashion Complementary Image Generation.}
Fashion CIG synthesizes garments that visually complement a given seed item, such as generating bottoms to pair with a top.
Early GAN-based methods~\cite{DBLP:journals/ijon/LiuSCM20,DBLP:conf/ijcai/LiuSRNT020,DBLP:journals/corr/abs-2408-09847} translated seed features into compatible targets but were limited in resolution, realism, and controllability.
Diffusion-based approaches, such as DiFashion~\cite{DBLP:conf/sigir/XuWFMZ024} and its variants~\cite{DBLP:conf/sigir/YuM00L025}, improve fidelity by conditioning on seed images and structured template prompts.
However, template prompts do not capture how users may express preferences through free-form language with varying specificity.
Our work reframes CIG as a multimodal language-grounding task by introducing free-form instructions and analyzing how instruction specificity affects generation and reliance on visual context.

\paragraph{Generative Models for Image Synthesis.}
Early fashion image generation approaches relied on GANs~\cite{DBLP:journals/corr/GoodfellowPMXWOCB14,DBLP:conf/icdm/KangFWM17,DBLP:conf/cvpr/KarrasLA19}, which achieved visual realism but often suffered from unstable training and limited diversity~\cite{DBLP:conf/icml/MeschederGN18}.
Latent Diffusion Models (LDMs)~\cite{DBLP:conf/nips/HoJA20,DBLP:conf/iclr/SongME21} improved fidelity and controllability, and have been widely adopted in fashion tasks such as virtual try-on~\cite{DBLP:conf/eccv/MorelliFCLCC22,DBLP:conf/mm/MorelliBCC0C23} and image editing~\cite{DBLP:conf/iccv/BaldratiMCC0C23,DBLP:conf/aaai/WangY24a,DBLP:conf/aaai/HouMM0025}.
More recently, Flow Matching (FM)~\cite{DBLP:conf/iclr/LipmanCBNL23,DBLP:conf/iclr/LiuG023} has emerged as an alternative that learns deterministic mappings through ODE-based probability paths, enabling stable training and faster sampling.
Despite these advantages, FM remains unexplored for fashion CIG.

\paragraph{Conditioning Strategies for Generative Models.}
Joint image-text conditioning has been extensively studied in image editing, where outputs usually preserve the spatial structure of the input.
Methods such as~\cite{DBLP:conf/iccv/ZhangRA23,DBLP:conf/aaai/MouWXW0QS24,DBLP:conf/cvpr/BrooksHE23} introduce control branches, adapters, or instruction-specific fine-tuning to inject visual conditions into text-to-image backbones.
However, complementary generation differs from editing because the target belongs to a different garment category and must remain stylistically compatible with the seed.
Recent CIG methods~\cite{DBLP:conf/sigir/XuWFMZ024,DBLP:conf/sigir/YuM00L025} address this by adding auxiliary conditioning modules to pretrained generative backbones.
In contrast, \styleflow integrates seed image and text conditioning in a single multimodal transformer, avoiding additional branches.

\section{Conclusion}
This work introduced a new formulation of CIG that conditions generation on a seed garment and a natural-language query. To support this setting, we enriched three CIG benchmarks with instructions at multiple specificity levels, enabling controlled evaluation of how language detail affects generation quality, catalog alignment, and visual grounding. We proposed \textbf{\styleflow}, a RFM model that jointly conditions on seed images and free-form instructions within a multimodal transformer, avoiding auxiliary conditioning branches while preserving textual intent and visual context. \styleflow produces realistic, compatible, and instruction-aligned garments. Future work will extend this setting beyond top--bottom pairs and incorporate real user queries, preferences, and interaction histories.

\bibliography{custom}

\appendix

\clearpage

\section{More Details On Flow Matching}\label{app:fm_more}
Flow Matching (FM)~\cite{DBLP:conf/iclr/LipmanCBNL23} is a recent generative modeling paradigm based on deterministic probability paths, offering an alternative to score-based diffusion models.
Given a target distribution $q$ in $\mathbb{R}^d$ and a simple base distribution $X_0$ (e.g., Gaussian), FM defines a family of intermediate distributions $\{p_t\}_{t \in [0,1]}$ that smoothly interpolate from $X_0$ to $X_1 = q$. This interpolation is governed by an ordinary differential equation (ODE):
\begin{equation}
\frac{d}{dt}\psi_t(x) = u_t(\psi_t(x)), \quad \psi_0(x) = x,
\end{equation}
where $u_t$ is a time-dependent velocity field parameterized by a neural network. Sampling is performed by integrating this ODE from noise to data. Thus, to learn a mapping between the known distribution $p_0$ and the data distribution $p_1$, it is sufficient to minimize the following loss function:
\begin{equation}
\mathcal{L}_{FM}(\theta) = \mathbb{E}_{t}\!\left[\|u_t^\theta(X_t)-u_t(X_t)\|^2\right].
\end{equation}

Since the exact velocity $u_t$ is unknown, training uses a surrogate objective. Conditional Flow Matching (CFM)~\cite{DBLP:conf/iclr/LiuG023,DBLP:conf/iclr/LipmanCBNL23} frames the problem by conditioning on pairs $(X_0, X_1)$ drawn from $(p_0, p_1)$ and defining an interpolation path:
\begin{equation}
X_t = \Psi(X_0, X_1),
\end{equation}
which leads to the tractable objective by conditioning from the target variable $X_1$:
\begin{equation}
\begin{aligned}
\mathcal{L}_{\mathrm{CFM}}(\theta)
&= \mathbb{E}_{t,X_t\mid X_1,X_1}\!\Big[ \\
&\quad \left\|u_t^\theta(X_t)-u_t(X_t\mid X_1)\right\|^2
\Big].
\end{aligned}
\end{equation}
Thus, CFM learns a velocity field consistent with a chosen path between the base and target distributions.

Rectified Flow Matching (RFM)~\cite{DBLP:conf/iclr/LiuG023,DBLP:conf/icml/EsserKBEMSLLSBP24} is a specific instantiation of CFM that enforces a \emph{rectified} path (e.g., a straight-line trajectory) between a known Gaussian prior and the data distribution:
\begin{equation}
X_t = t X_1 + (1-t)X_0, \quad X_0 \sim \mathcal{N}(0,I).
\end{equation}
This choice simplifies the approximation of the velocity field by reducing the objective curvature.

\paragraph{Comparison to Diffusion Models.} Score-based Latent Diffusion Models (LDMs)~\cite{DBLP:conf/nips/HoJA20,DBLP:conf/iclr/0011SKKEP21} approach generation by simulating a stochastic differential equation (SDE) that progressively denoises an initial Gaussian sample, estimating the score function. This process requires iterative evaluations of the generative backbone to approximate the reverse diffusion process. Sampling cost therefore remains an important design consideration for interactive settings such as fashion recommendation.

In contrast, RFM reduces curvature in the probability path and simplifies the velocity field, which can support stable generation with a moderate number of integration steps.
This is relevant in the context of fashion complementary item generation, where generation quality and sampling budget both matter for interactive systems. In our experiments, \textit{StyleFlow} uses a 20-step sampling schedule while maintaining strong fidelity and instruction alignment. We use these measurements as practical context rather than as the primary contribution of the paper.

\section{More Examples on Free-Form Instruction Synthesis}\label{app:data_constr}

\begin{figure}[t!]
    \centering
    \begin{tcolorbox}[width=\linewidth, colback=gray!5, colframe=gray, boxrule=0.5pt, title=Prompt for First Stage]
    You are given an image of a bottom garment. Analyze its color, fit, material, type, and style. Then, generate a natural-sounding search query that a user might type into an image search engine to find this kind of bottom garment. Make the query short, descriptive, and specific -- as if someone is looking for this item online. Respond only with the query.
    \end{tcolorbox}

    \begin{tcolorbox}[width=\linewidth, colback=gray!5, colframe=gray, boxrule=0.5pt, title=Prompt for Second Stage]
    You are writing prompts for a generative image model focused on fashion garments, not people.

    Rewrite the following product description into three prompts separated by semicolons:\\

    1. \textit{Detailed}: Generate an image of the garment with full description including type, color, fabric, fit, and details.\\
    2. \textit{Medium}: Generate an image of the garment including type, fit, optionally color; keep concise. \\
    3. \textit{Minimal}: Generate an image of the garment mentioning only type, optionally one simple detail; very brief.\\

    Example input: "black cotton jogger sweatpants with an elastic waistband"
    Example output:
    An image of black cotton jogger sweatpants with an elastic waistband and ribbed cuffs; Generate an image of black tapered jogger sweatpants; Generate an image of jogger pants.

    \end{tcolorbox}

    \caption{Examples of OpenAI-o3 prompts used to generate free-form instructions at different levels of detail, as described in Section 2.2 of the paper.}
    \label{fig:prompts}
\end{figure}

\begin{figure*}[t!]
    \centering
    \includegraphics[width=\linewidth]{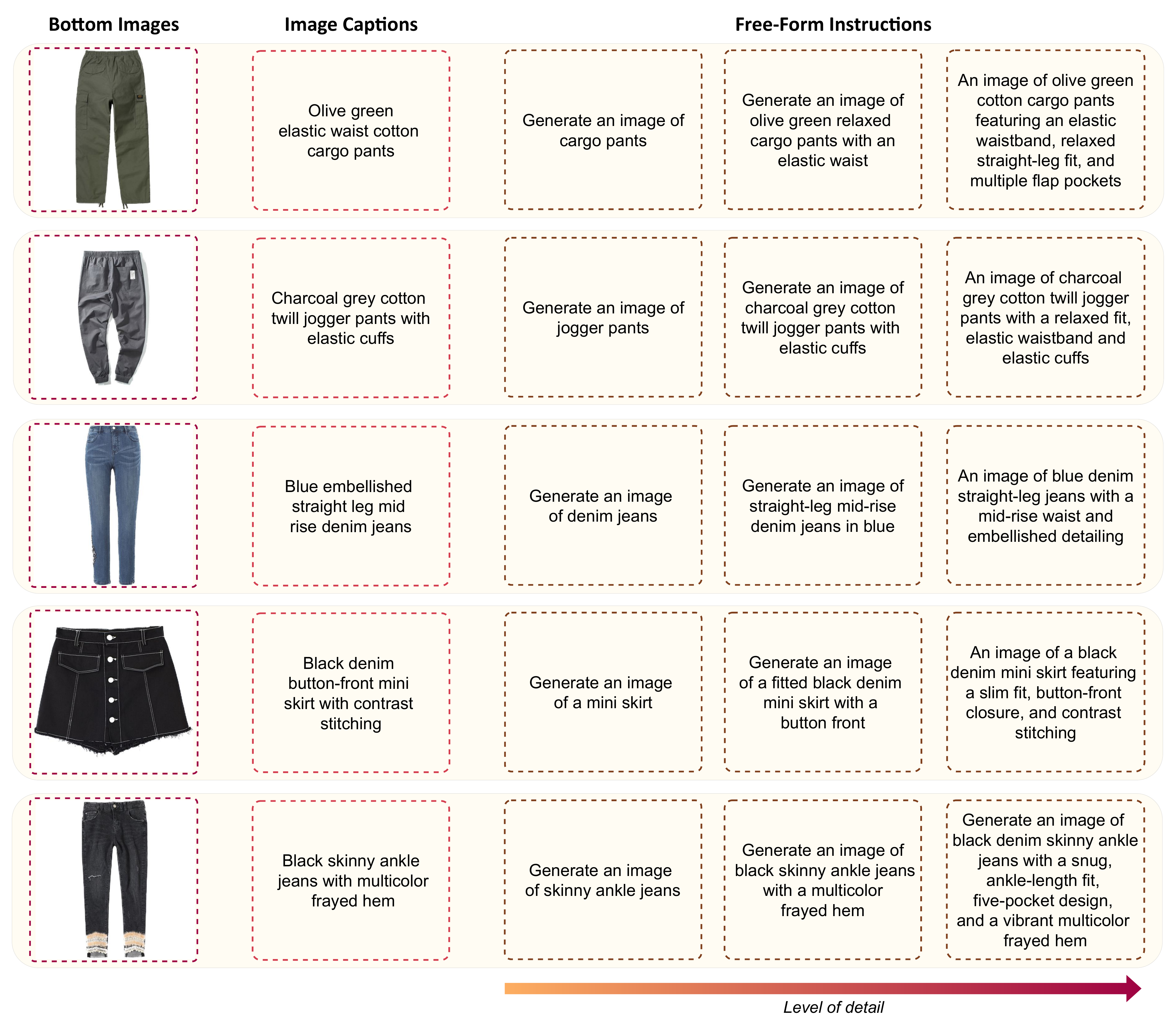}
    \caption{Examples of generated captions and free-form instructions: OpenAI-o3 first generates a caption from the query image, which is then used to produce three free-form instructions.}
    \label{fig:prompt_construction}
\end{figure*}

\begin{figure*}[t!]
    \centering
    \includegraphics[width=0.98\linewidth]{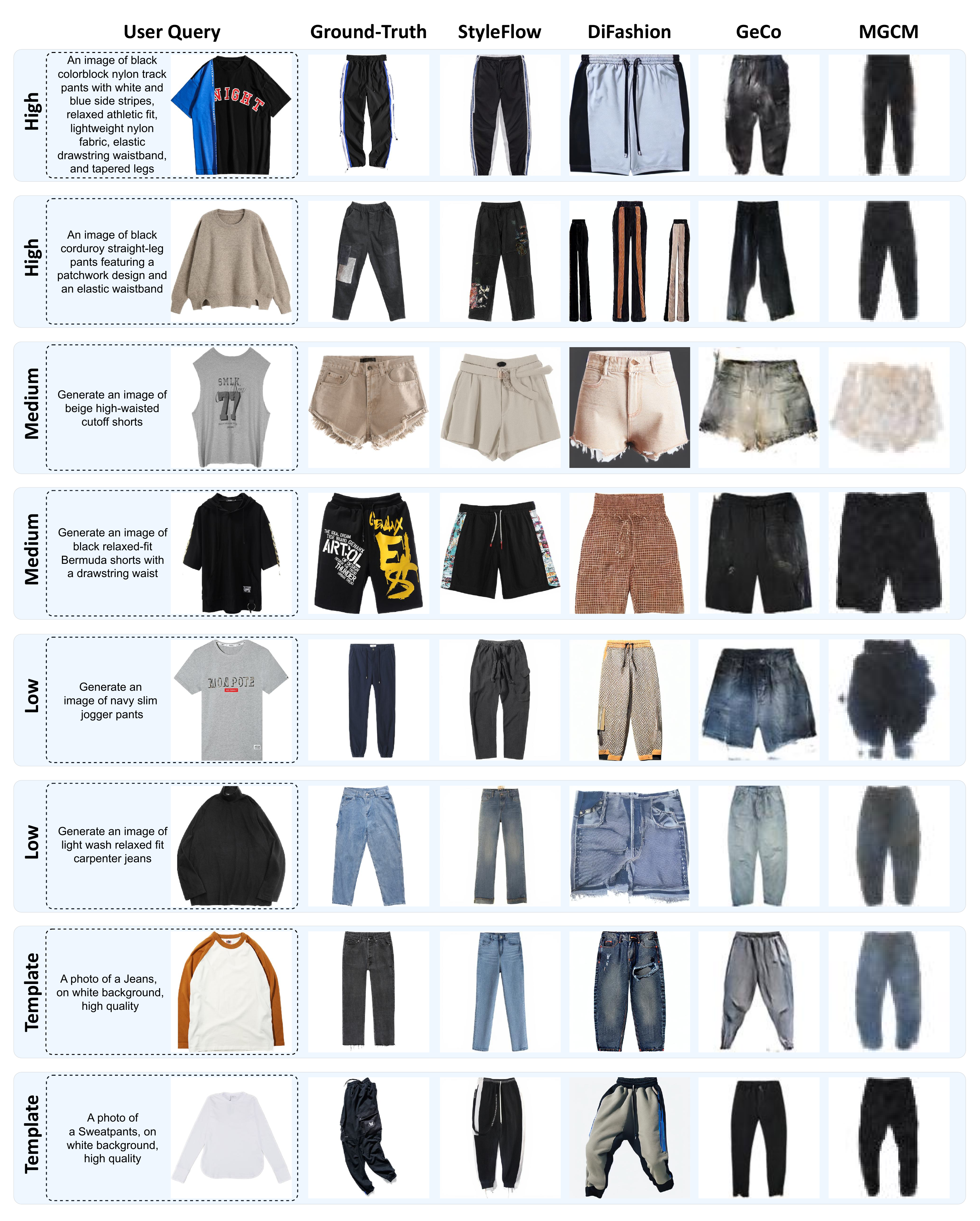}
    \caption{Additional samples of bottom garment generation from user queries composed of a top image and instructions at
varying levels of detail.}
    \label{fig:gen_appendix}
\end{figure*}

As shown in~\cref{fig:prompts}, the enrichment pipeline relies on carefully designed prompting strategies to obtain diverse yet realistic textual annotations.
The first stage prompts the model to analyze visual attributes such as color, fit, material, type, and style, returning a caption for each analyzed image.
The second stage takes these captions and reformulates them into three levels of natural language free-form instructions: \textit{high}, \textit{medium}, and \textit{low} detail. This progressive reduction in attribute coverage simulates varying levels of query specificity, from detailed to generic queries. Together, these prompts provide semantically rich and stylistically grounded instructions for controlled evaluation across different prompt granularities.

\cref{fig:prompt_construction} depicts some examples of generated captions and free-form instructions. All generated instructions underwent full manual verification to ensure correctness, fluency, and relevance to the garment image. In cases where the model output was incomplete, ambiguous, or inconsistent with the visual content, instructions were revised or regenerated.
This enrichment process was applied uniformly across all datasets (FashionVC, ExpReduced, and FashionTaobao‑TB), ensuring consistent evaluation conditions. Test splits were constructed so that bottom garments in the test set did not appear in training, even under different pairings, preventing leakage.
The resulting benchmark thus provides a rich set of user‑like queries that enable controlled experimentation on how instruction detail affects CIG.

\section{Human Evaluation Protocol Details}\label{app:human_eval_details}

Both human studies were administered through institution-hosted web interfaces. Participants were recruited among researchers. The landing page informed participants that the study had no commercial purpose and that the collected information would be handled in compliance with applicable privacy regulations. Participants proceeded only after reading this information and pressing the \emph{Start} button, which served as informed consent to take part in the study.

The backend stored only a random session identifier, browser user agent, screen resolution, the sampled items shown in the trial, and the submitted ratings. No direct identifiers such as names or email addresses were collected. Each participant completed 8 trials.

\paragraph{Instruction Evaluation Study.}
Each trial displayed one reference garment image and three automatically generated descriptions associated with that garment. Participants rated each description by answering: ``How would you evaluate this text?'' and ``Do you think this instruction is aligned with the image?'' The first question was answered on an ordinal scale with half-step increments; in the paper, we summarize this dimension as perceived instruction quality in terms of clarity, fluency, and informativeness. The second question was answered as a binary yes/no judgment. The order of items within each trial was randomized.

\paragraph{Generation Evaluation Study.}
Each trial displayed one top garment and three candidate bottom garments. When more than three candidates were available for the selected top image, the interface sampled three of them uniformly at random. Participants evaluated each candidate by answering: ``How do you rate the visual quality of the image?'', ``How compatible do you think the top and bottom are as a pair?'', and ``Do you think that this image is generated?'' The first two questions were answered on ordinal rating scales with half-step increments, while the last question was answered as a binary yes/no judgment. Candidate order was randomized in each trial.

\paragraph{Interface Description.}
The landing page for the instruction study stated that participants would see a reference image together with three automatically generated textual descriptions and would judge description quality and adherence to the image. The landing page for the generation study stated that participants would see a top garment together with recommended bottom garments and would judge visual quality and compatibility. In the generation study, the authenticity question was additionally asked within each trial, as described above.

\section{Additional Implementation Details}\label{app:additional_imp_details}
To support reproducibility, we provide further implementation details.
\paragraph{Datasets and Splits.}
Experiments are conducted on three established fashion compatibility benchmarks: \textit{FashionVC}~\cite{DBLP:conf/mm/SongFLLNM17}, \textit{ExpFashion}~\cite{DBLP:journals/tkde/LinRCRMR20}, and \textit{\fttb}~\cite{DBLP:journals/corr/abs-2408-09847}, all containing annotated top--bottom garment pairs.
Following common CIG protocols~\cite{DBLP:journals/ijon/LiuSCM20,DBLP:conf/ijcai/LiuSRNT020}, we use a reduced split of ExpFashion, denoted as \textit{ExpReduced}, to enable controlled comparison across benchmarks with comparable pair distributions.
FashionVC includes 18,640 pairs at \(128{\times}128\) resolution, ExpReduced includes 18,640 pairs at \(224{\times}224\), and \fttb includes 88,326 pairs at \(512{\times}512\).
All datasets are enriched with free-form natural-language instructions as described in the main paper.
The evaluation focuses on top--bottom pairings, which align with prior CIG work, and target bottom garments are disjoint across train, validation, and test splits.

\paragraph{Metric Computation.}
Generation quality is evaluated with FID~\cite{DBLP:conf/nips/HeuselRUNH17}, KID~\cite{DBLP:conf/iclr/BinkowskiSAG18}, LPIPS~\cite{DBLP:conf/cvpr/ZhangIESW18}, and CLIP-Score~\cite{DBLP:conf/emnlp/HesselHFBC21}.
FID and KID are computed from InceptionV3~\cite{DBLP:conf/cvpr/SzegedyVISW16} features, with all images resized to \(299 \times 299\).
LPIPS and CLIP-Score are computed at each model's native resolution, as both metrics internally resize inputs to \(224 \times 224\).
For catalog alignment, generated images and catalog items are embedded with a fixed SigLIP encoder~\cite{DBLP:conf/iccv/ZhaiM0B23}; catalog items are ranked by cosine similarity, and alignment is measured using MRR, Recall, and nDCG.

\paragraph{Baselines.}
DiFashion~\cite{DBLP:conf/sigir/XuWFMZ024} is adapted to condition on both the top garment and free-form instructions while preserving its original training protocol.
For DiFashion, we followed the original implementation and set the hidden dimension to 256 to maintain compatibility with the pretrained weights. The category guidance scale is set to 3.5, as explored in the original paper and aligned with the evaluation setup used for \styleflow, while the mutual guidance scale is fixed at 5, in line with the best-performing configuration reported by the authors.
GeCo~\cite{DBLP:journals/corr/abs-2408-09847} and MGCM~\cite{DBLP:journals/ijon/LiuSCM20} are extended with a pretrained CLIP text encoder~\cite{DBLP:conf/icml/RadfordKHRGASAM21} to support free-form instruction conditioning.
For all baselines, hyperparameters were explored within the ranges specified in their respective papers, and the best-performing configuration on the validation set was selected.

\paragraph{Training Setup.}
\styleflow and DiFashion were trained and evaluated with a fixed random seed of 0, while the remaining baselines used a seed of 42, consistent with their original implementations.
For \styleflow, training uses a batch size of 1, gradient accumulation over 8 steps, bfloat16 precision, prompt dropout with probability 0.1, and classifier-free guidance with scale 3.5 at inference.
All experiments are conducted on a single NVIDIA H100 GPU.
The MM-DiT input channels are extended to incorporate the seed-image latent representation, and sampling is performed with the Euler ODE solver using 20 integration steps unless otherwise stated.
\section{More Examples of Generations}\label{app:more_gen_examples}

\cref{fig:gen_appendix} provides extended qualitative results across all prompt granularities, including \textit{low}, \textit{medium}, \textit{high}, and \textit{template}-based settings.
\styleflow consistently synthesizes visually coherent, realistic bottom garments that align with the reference top and the given instruction. In detailed and medium settings, where the instructions provide rich semantic cues, the model performs particularly well, accurately capturing fine‑grained attributes such as color, fit, and style.

By contrast, under the \textit{template} setting, where prompts follow a structured template with minimal detail, the behavior of \styleflow is closer to the \textit{low} setting, as both provide limited guidance. In these cases, the outputs remain stylistically compatible but show reduced specificity compared to the richer prompt settings.

DiFashion exhibits limited variation across prompt granularities, often generating outputs that are visually similar regardless of whether the instructions are detailed or minimal.
GAN‑based baselines often yield blurrier or less plausible outputs, and DiFashion occasionally introduces irrelevant elements even in detailed settings. Despite these challenges, \styleflow maintains strong performance and robustness across all prompt types.

\end{document}